\documentclass[letterpaper, 10 pt, conference]{ieeeconf}
\makeatletter
\let\MYcaption\@makecaption
\makeatother

\usepackage[font=footnotesize]{subcaption}

\makeatletter
\let\@makecaption\MYcaption
\makeatother

\usepackage{graphics} 
\usepackage{epsfig} 
\usepackage{amsmath} 
\usepackage[T1]{fontenc}
\usepackage{bm}
\usepackage{cite}
\usepackage{here}
\usepackage{booktabs}
\usepackage{threeparttable}
\usepackage{multirow}
\usepackage{color} 
\usepackage{cases}
\usepackage{siunitx}

\usepackage{algorithm}
\usepackage{algpseudocode}
\usepackage{makecell}
\usepackage{outlines}
\usepackage{amssymb}

\usepackage{float}

\usepackage{url}
\usepackage{tabularray}
\usepackage[normalem]{ulem}
\usepackage{mathrsfs}
\usepackage{multirow}
\usepackage{tabularray}

\IEEEoverridecommandlockouts                              
\title{\LARGE \bf 
UltraDP: Generalizable Carotid Ultrasound Scanning with Force-Aware Diffusion Policy}
\author{Ruoqu Chen, Xiangjie Yan, Kangchen Lv, Gao Huang, Zheng Li, and Xiang Li
\thanks{R. Chen, X. Yan, K. Lv, G. Huang, and X. Li are with the Department of Automation, Tsinghua University. Zheng Li is with the Department of Surgery, Chow Yuk Ho Technology Centre for Innovative Medicine, Li Ka Shing Institute of Health Science and Multi-scale Medical Robotics Center, The Chinese University of Hong Kong, Hong Kong.
This work was supported in part by the National Key R\&D Program of China under Grant 2024YFB4708200, in part by Tsinghua University Initiative Scientific Research Program, in part by the National Natural Science Foundation of China under Grant U21A20517 and 62461160307, and in part by the BNRist project under Grant BNR2024TD03003.}}

\begin{document}
\maketitle
\pagestyle{empty}  
\thispagestyle{empty} 

\begin{abstract}
Ultrasound scanning is a critical imaging technique for real-time, non-invasive diagnostics. However, variations in patient anatomy and complex human-in-the-loop interactions pose significant challenges for autonomous robotic scanning. Existing ultrasound scanning robots are commonly limited to relatively low generalization and inefficient data utilization. To overcome these limitations, 
we present UltraDP, a Diffusion-Policy-based method that receives multi-sensory inputs (ultrasound images, wrist camera images, contact wrench, and probe pose) and generates actions that are fit for multi-modal action distributions in autonomous ultrasound scanning of carotid artery. We propose a specialized guidance module to enable the policy to output actions that center the artery in ultrasound images. To ensure stable contact and safe interaction between the robot and the human subject, a hybrid force-impedance controller is utilized to drive the robot to track such trajectories. Also, we have built a large-scale training dataset for carotid scanning comprising 210 scans with 460k sample pairs from 21 volunteers of both genders. By exploring our guidance module and DP's strong generalization ability, UltraDP achieves a 95$\%$ success rate in transverse scanning on previously unseen subjects, demonstrating its effectiveness.
\end{abstract}

\section{Introduction}
Ultrasound scanning is a crucial imaging method that allows real-time, radiation-free, non-invasive, and cost-effective evaluation of multiple organs, such as the carotid artery, heart, and liver, to support clinical diagnostics. 
It is a high-cognitive-burden task, as it requires a trained sonographer (training can last up to several years) with specialized knowledge of the specific organ—something that can be challenging to fulfill in medically underserved areas. Additionally, it is a high-intensity, repetitive job \cite{harrison2015work} for sonographers due to the growing demand for ultrasound diagnostics. These two factors have brought robotic ultrasound scanning to the forefront of discussion.
\begin{figure}[!t]
    \centering
    \includegraphics[width=0.8\linewidth]{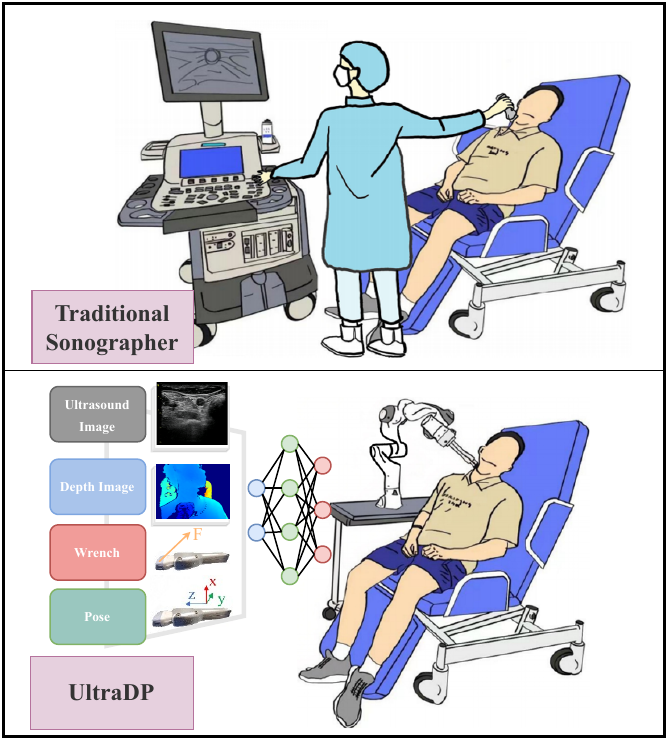}
    \vspace{-1mm}
    \caption{Demonstration of carotid artery ultrasound scanning task. Top: traditional sonographer. Bottom: the ultrasound robot with the proposed method UltraDP, which takes ultrasound images, depth images, contact wrench, and probe pose as input and outputs the desired pose and contact wrench.}
    \label{fig_head_figure}
\end{figure}

One of the key performance indicators of ultrasound robots is generalization. In the context of ultrasound, generalization refers to the ability of the robot to adapt and perform accurately across a wide range of different patient conditions and anatomical variations without requiring extensive reprogramming or adjustments. This capability is crucial because it ensures that ultrasound robots can function effectively in real-world clinical settings where patients present with diverse conditions and characteristics. 

Achieving generalization in ultrasound robots is challenging because it demands the navigation system to comprehend organ anatomy, interpret ultrasound images, and possess spatial imagination skills to map 2D ultrasound images onto 3D organ structures. Each probe adjustment made by sonographers during a scan is, in fact, the outcome of the integrated application of these specialized knowledge and skills. 
Researchers tried to depict the ``knowledge and skills'' using well-designed rules\cite{10715958,yan2024unified,huang2018fully,10684288}, which was called rule-based methods, usually with a specialized perception system. However, despite meticulous designs, the generalization ability remained limited, especially when scaling to a larger number of patients with diverse body types. For example, a set of parameters suitable for an overweight male may not work for a slim female.

Another attempt to improve generalization ability is learning-based methods like imitation learning \cite{si2023unified,deng2021learning,10878457,jiang2024cardiac,jiang2025structureaware,jiang2024sequenceaware,jiang-2025}. It has gained popularity for discovering the underutilized value in thousands of routine ultrasound scans.
However, existing works did not consider the multi-modal action distributions featured by ultrasound scanning tasks: if the ultrasound image is blurred or contains significant portions lacking visibility (e.g., appearing black), multiple corrective actions can be taken to improve image quality, such as changing probe directions, translations, or adjusting the contact force/torque. Hence, the generalization ability could fall short on real-world scanning tasks.

Recently, Diffusion Policy (DP) has shown strong potential in learning multi-modal action distributions and generating smooth trajectories in robotic manipulation tasks \cite{chi2023diffusion,ze20243d}. This paper takes advantage of DP and offers new possibilities for generating actions that align with the essence of the ultrasound task. Such an application is not trivial because the target of ultrasound scanning is different from other tasks achieved by DP (for example, ensuring a clear ultrasound image and keeping the carotid artery in the center of the image in the carotid scanning task) and cannot be learned without any modification. Also, the subject is a dynamic human, posing new implementation challenges. 

To achieve this, we propose UltraDP, a learning-based method for transverse section ultrasound scanning of carotid artery that mimics human sonographers' hand-eye coordination and takes ultrasound images, wrist camera images, contact wrenches, and probe poses as multi-modal inputs and predicts the desired pose and contact wrench for navigation; at the lower control level, it employs a hybrid force-impedance controller. This paper makes the following contributions to integrate DP into ultrasound systems and hence strengthening its generalization ability:
\begin{enumerate}
\item We constructed a large-scale dataset of expert sonographers' demonstrations, including ultrasound images, wrist camera images, probe poses, and contact wrenches. The dataset comprises 210 scans with 460k sample pairs collected from 21 volunteers of both genders. The dataset will be published after careful processing to prevent information leakage.
\item We designed a specialized guidance module in diffusion policy for the transverse carotid ultrasound scanning task to enable UltraDP to keep the artery horizontally centered in the ultrasound image, which is an important requirement for carotid scanning.
\item We conducted extensive real-world experiments on previously unseen volunteers and showed UltraDP’s superior performance and better generalization ability.
\end{enumerate}

We trained our policy with real-world data from 210 scans (460k sample pairs) collected from 21 volunteers by two certified sonographers, and collected 54k sample pairs from new male and female volunteers to further validate its generalization ability. 
We conducted experiments comparing UltraDP with a rule-based Visual servoing method \cite{yan2024unified} and the classic Behavior Cloning method. We also did ablation studies for the sensory modalities and a subjective study. The results showed better generalization ability, the necessity of the modalities, and improved comfort for subjects using UltraDP.
    
\begin{figure}
    \centering
  \subcaptionbox{}
    {\includegraphics[width=0.39\linewidth]{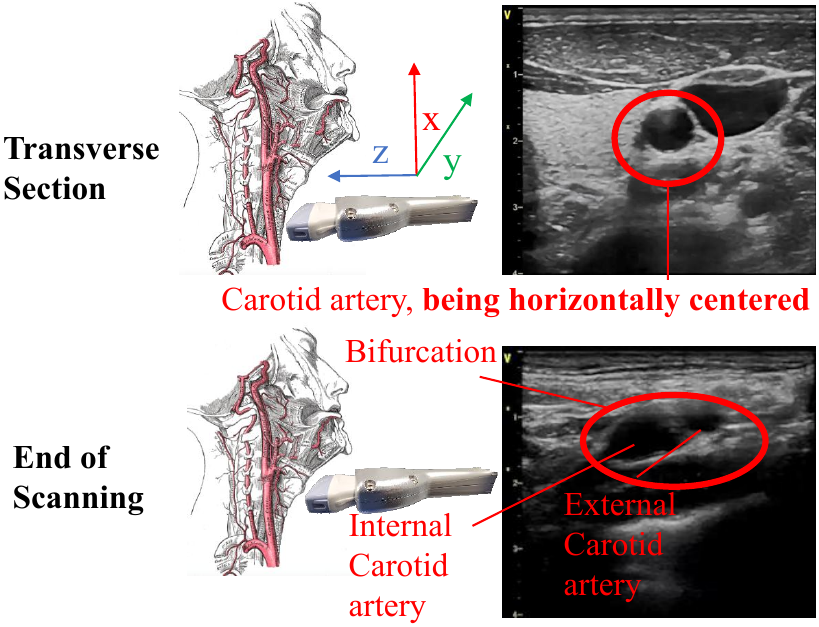}}   
    \subcaptionbox{}{
    \includegraphics[width=0.58\linewidth]{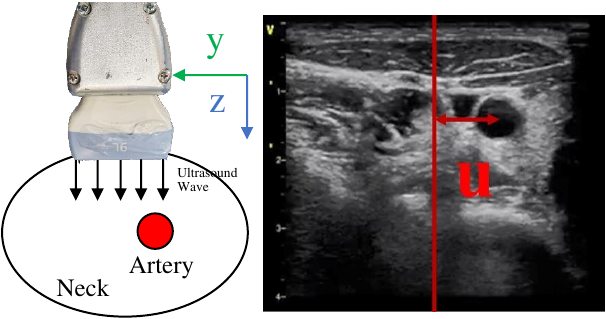}}
    
    \caption{(a) Probe positions and the corresponding ultrasound images for transverse section and at the end of scanning. (b) Illustration of ultrasound imaging principle.}
    \label{fig_carotid}
\end{figure}

Before proceeding, let’s delve into the requirements of the carotid artery transverse section ultrasound scanning task (see Fig. \ref{fig_head_figure} and Fig. \ref{fig_carotid}(a)). Carotid artery ultrasound scanning is typically performed to check for plaque build-up or blood clots that could restrict blood flow and potentially lead to a stroke\cite{mayo}. The task procedure unfolds as follows: a sonographer, holding an ultrasound probe, begins at the lower part of the neck and moves upwards along its length to capture transverse-section ultrasound images (see Fig. \ref{fig_carotid}(a)), upon identifying the bifurcation of the internal and external carotid arteries. Throughout the procedure, an important requirement is to keep the artery centered horizontally in the image so that during subsequent longitudinal scanning, the longitudinal section of the artery can be more easily located.

\section{Related works}
\subsection{Rule-Based Ultrasound Systems}
Many existing works rely on rule-based navigation \cite{duan2022ultrasound,goel2022autonomous,huang2021towards,yan2024unified}. Researchers achieved scans by setting simple rules like keeping the anatomic landmark in the center of the ultrasound image \cite{10715958}, or designing a finite-state machine for different scanning modes \cite{yan2024unified}. These methods are easy to implement but lack generalization to unforeseen situations. For example, if the perception system fails to recognize an anatomical landmark \cite{10715958}, the entire system breaks down. More commonly, when model parameters — such as neck length, artery angle, and other offsets described in \cite{yan2023multi} — do not fit a new patient, the resulting ultrasound images are consistently poor.
\subsection{Learning-Based Ultrasound Systems}
As examples of supervised learning, in \cite{jiang2024cardiac,jiang2025structureaware}, authors trained a set of networks, including Transformer and ResNet, to form a world model for cardiac scanning; researchers in \cite{10802113} trained a self-supervised network based on synthetic data from simulations. 
These methods trained huge networks for robots with large-scale clinical data but gained limited ability to generalize.
DP is a state-of-the-art imitation learning method boasting the multi-modal action representation ability and has achieved success on some contact-rich manipulation tasks\cite{liu2024forcemimic,wu2024tacdiffusion,hou2024adaptive}. In \cite{liu2024forcemimic}, authors achieved force-aware imitation learning using DP with a lower-level hybrid force-position controller to do the vegetable peeling task. A similar solution was used to do the peg-in-hole task in \cite{wu2024tacdiffusion}. In \cite{hou2024adaptive}, force data was processed by Fast Fourier Transform before input into diffusion policy, and the action is not the desired force but stiffness and virtual target.
However, DP has not been validated in ultrasound scanning tasks.

\begin{figure*}[h]
    \centering
    \includegraphics[width=1.0\linewidth]{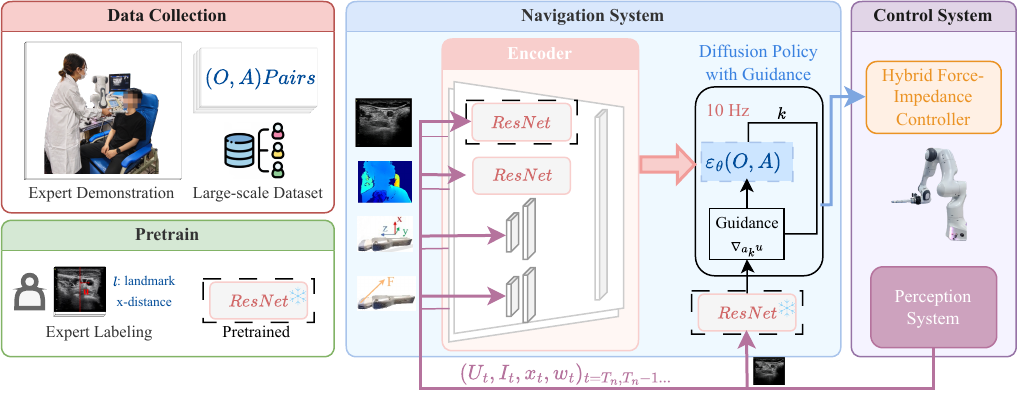}
    \caption{The structure of UltraDP, including data collection, pretrain, navigation system and control system. In the navigation system module, only the inference process is illustrated, where the ``ResNet'' block within the dashed box means the trained network on the pretrained parameters, and the ``ResNet'' block with a ``frozen'' sign within the dashed box equals the one in the pretrain module.}
    \label{structure1}
\end{figure*}

Compared to previous works on DP-based contact-rich manipulation tasks \cite{liu2024forcemimic,wu2024tacdiffusion,hou2024adaptive}, our ultrasound scanning task presents a significantly greater challenge for DP to learn. This is because (\romannumeral1) it involves a dynamic human subject rather than a static environment. The contact wrench and the action ranges vary among different people, causing difficulties in the training of the policy (\romannumeral2). It requires the policy to interpret ultrasound images, which means the task is not just in cartesian space (trajectory-based or goal-based) but a ``higher-level-goal-based'' task, which is difficult to learn.

\section{Methodology}

The proposed method utilized a modified diffusion policy to learn the desired wrench and pose from sonographer demonstrations. The structure of the UltraDP system is shown in Fig. \ref{structure1}. 
There are four main modules: Data collection, Pretrain, Navigation system, and Control system, which will be discussed in the following sections.

\subsection{Data Collection}
We collected real-world demonstration data by recording the ultrasound images \(\bm U\in \mathbb{R}^{H_1\times W_1}\), the probe poses \(\bm x\in SE(3)\), the contact wrench \(\bm w\in \mathbb{R}^6\), and the RGBD images \(\bm I\in\mathbb{R}^{H_1\times W_1\times 4}\) from the wrist camera while a sonographer performed a complete scan, holding the manipulator flange where the force sensor was mounted. We collected multiple demonstration trajectories for each volunteer, ensuring slight variations in position and posture across different trajectories from the same volunteer. Thus, we got a demonstration trajectory dataset:
\begin{equation}
    \mathcal{D}=\{d_1,d_2,\cdots,d_n|d_i = (\bm U_t, \bm I_t,\bm x_t,\bm w_t)_{t=1:T_n}\}.
\end{equation}
Then we formed it into observation-action pairs:
\begin{align}
\bm O_t &= (\bm U_t, \bm I_t,\bm x_t,\bm w_t)\\
\bm A_t &= \left(diff(\bm x_{t+1},\bm x_t),\bm w_t\right),
\end{align}
where \(diff()\) calculated the relative pose from \(\bm x_{i+1}\) to \(\bm x_{i}\).
We did data augmentation that each demonstration trajectory was diffused by several random transformations noise in cartesian space.
While preprocessing the RGBD images \(\bm I\) from the wrist camera, a mask was generated according to the depth channel, removing the irrelevant background. 
The end effector pose was represented using a 3D Cartesian position and a 6D rotation representation as proposed in \cite{zhou2019continuity}. The observation and action pair sequences \(\{(\bm{O}, \bm{A})\}\) are then taken by the model for training with a receding horizon.

During data collection, our hybrid force-impedance interaction controller was activated with all stiffness and damping parameters set to zero, except for the force control, which assisted the sonographer in establishing contact effortlessly while remaining fully compliant with his or her movements along the skin surface. 
This support enables the sonographer to complete a high-quality scan in approximately 30 seconds — less than half the time required without assistance. 
The entire data collection process was approved and supervised by the University Science and Technology Ethics Committee.

Note that these demonstration data were collected by experts with proficient ultrasound scanning experience. The anatomical landmark positions in the images in the demonstrations were always the center. It is rather difficult for DP to learn automatically the centering movement during the training. So we designed a guidance step, which will be discussed later in Sec \ref{sec_method_3}.

\subsection{Pretrain}
The encoder for Ultrasound images \(\bm{U}\) (implemented as a ResNet in our approach) was pretrained on a classification-and-regression task using a small ultrasound image dataset including 6k pairs of\((\bm{U},l)\), where \(l\in \mathbb{R}\) denotes the position of the artery landmark along the \(x\)-dimension of the image. If the carotid artery is not present in the image, the classification head outputs 0 and the regression head outputs invalid values. If it is present, the classification head outputs 1, and the regression head predicts the position of the carotid artery \(l\in[0,1]\).
The ground truth of the dataset was labeled by expert sonographers.

After pretraining, the fully connected layers were removed, allowing the encoder—now enriched with artery landmark information—to serve as the initialization for subsequent training in the downstream encoder layer in the DP part.

Furthermore, the pretrained ResNet was also employed as a module to output the \(x\)-dimension coordinate of the carotid artery (measured in pixels) within the Navigation system. 

\begin{figure}[!t]
    \centering
    \includegraphics[width=\linewidth]{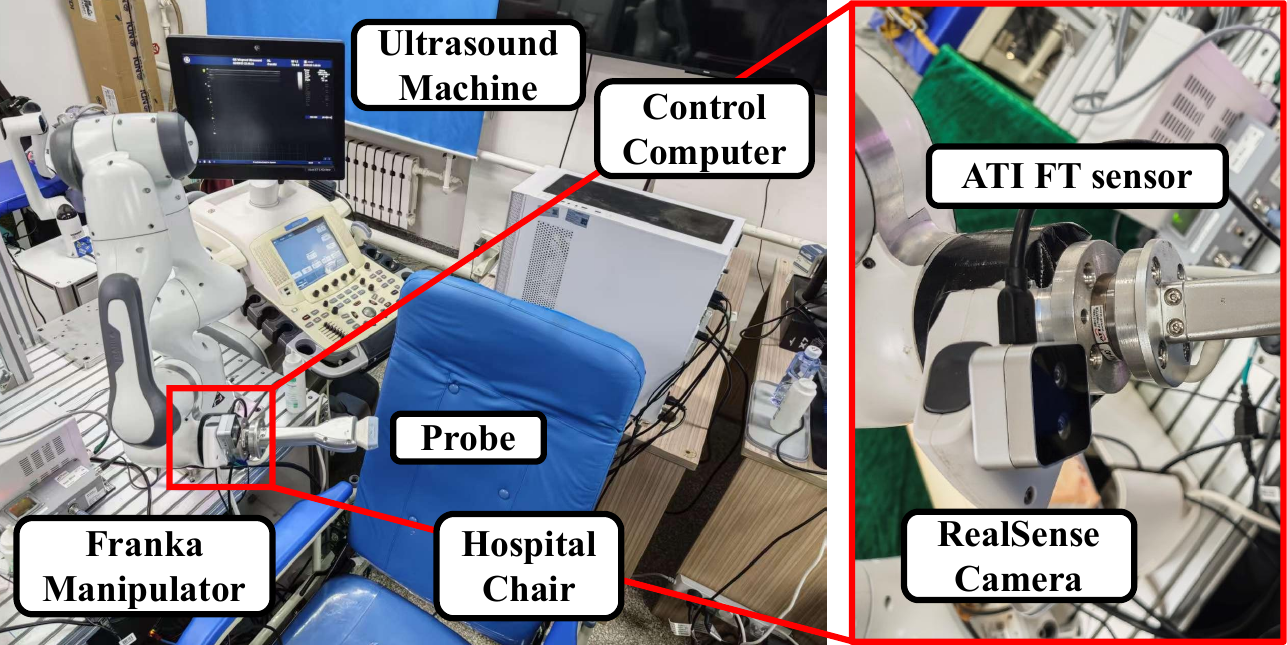}
    \caption{Demonstration of the experiment setup, which consists of a Franka manipulator, an ultrasound machine with a probe, an ATI mini 40 force/torque sensor connected between the arm flange and the ultrasound probe, and a RealSense D405 camera mounted on the arm flange.}
    \label{fig_exp_setup}
\end{figure}

\subsection{Navigation System for Ultrasound Scanning}\label{sec_method_3}
The main part of the navigation is a diffusion model that represents the robot's visuomotor policy using Denoising Diffusion Probabilistic Models (DDPM)\cite{chi2023diffusion, ho2020denoising}, which generates output actions through a denoising process as follows
\begin{equation}
\bm{a}^{k-1} = \bm \alpha ( \bm{a}^{k} -  \gamma \bm\varepsilon_{\theta}(\bm o,\bm{a}^{k},k) + \mathcal{N}(\bm 0,\sigma^{2}\bm I)),\label{eq_denoise}
\end{equation}
where $k$ denotes the steps of the denoising process, $ \mathcal{N}(0,\sigma^{2}\bm I))$ represents the Gaussian noise introduced at each iteration of the process, and $\varepsilon_{\theta}$ is the noise prediction network. Notice that here \(\bm o\) denotes the observation.


As mentioned earlier, the centering movement from expert sonographers is hard for DP to learn. Hence, during the inference process, the original denoising step (\ref{eq_denoise}) is added by a guidance term \(\bm g^k = \nabla_{\bm a^k} u\):
\begin{equation}
\bm{a}^{k-1} = \bm \alpha ( \bm{a}^{k} + \rho \bm g^k-  \gamma \bm\varepsilon_{\theta}(\bm o,\bm{a}^{k},k) + \mathcal{N}(\bm 0,\sigma^{2}\bm I)),\label{eq_denoise_add_guidance}
\end{equation}
where \(\rho\) is a scaling factor, and the gradient \(\nabla_{\bm a^k} u\) can be obtained by the imaging principle of ultrasound. For a linear array probe that performs a straight-line scan (see Fig. \ref{fig_carotid}(b)), the mapping is linear:
\begin{equation}
    \Delta y = a\Delta u, \label{eq_imaging_principle}
\end{equation}
where \(a\) is a known probe parameter, \(\Delta y\) is a displacement of the probe in cartesian \(y\) axis, \(\Delta u\) is the difference of pixel position of the artery. From (\ref{eq_imaging_principle}), we can get a constant gradient. Hence the guidance made every sample towards the position centering the carotid in the ultrasound image.

\subsection{Hybrid Force Impedance Interaction Control}
The dynamic model of the ultrasound scanning robot can be described as
\begin{eqnarray}
&\bm M(\bm q)\ddot{\bm q}+\bm C(\dot{\bm q}, \bm q)\dot{\bm q}+\bm g(\bm q)=\bm \tau+\bm \tau_{e},\label{eq_dyn}
\end{eqnarray}
where \(\bm M(\bm q)\hspace{-0.05cm}\in\hspace{-0.05cm}\Re^{n\times n},\bm C(\dot{\bm q}, \bm q)\dot{\bm q}\hspace{-0.05cm}\in\hspace{-0.05cm}\Re^n,\bm g(\bm q)\hspace{-0.05cm}\in\hspace{-0.05cm}\Re^n\) denote the mass matrix,  Coriolis and centrifugal term, and gravity vector respectively, and \(\bm \tau, \bm \tau_e\hspace{-0.05cm}\in\hspace{-0.05cm}\Re^n\) represent the control torque and the external torque that applied by the environment to the end effector respectively. In our setting, a Franka robot was used, so \(n=7\). Then, to ensure a safe interaction, the controller is proposed as
\begin{align}
    \bm \tau= \bm \tau_m+ \bm \tau_n+\bm g(\bm q)+\bm C(\dot{\bm q}, \bm q)\dot{\bm q},\label{eq_controller}
\end{align}
where \(\bm \tau_m\) is the control torque for the main scanning task, \(\bm \tau_n\) is that for the null-space task:
\begin{gather}
    \bm \tau_m = \bm{J}^T \bm{S}_i\left(-\bm K_{1}\tilde{\bm{x}}-\bm D_1 \dot{\bm{x}}\right)+\nonumber\\
    \bm{J}^T\bm S_f(\bm F_d+\bm K_f(\bm F_d-\bm F_{e})),\label{eq_tau1_regu}\\
    \bm \tau_n = (\bm I-\bm J^T\bm J^{\dagger T})\left(-\bm K_{2} \tilde{\bm{q}}-\bm D_2 \dot{\bm q}\right ) . \label{eq_tau2_regu}
\end{gather}
In (\ref{eq_tau1_regu}), \(\bm J\) is the Jacobian matrix, \(\bm S_i\) and \(\bm S_f\) are selection matrix for impedance control and force control, \(\tilde{\bm x}\) is the cartesian pose error \(\tilde{\bm x} = \bm x-\bm x_d\), \(\bm F_d\in \mathbb{R}^6\) is the desired contact wrench given by navigation module, \(\bm F_e\) is the external wrench measured by the wristed-mounted force sensor.

To maintain a steady and comfortable contact between the probe and the patient's neck, the force along the \(z\) axis is assigned to be force controlled (see Fig. \ref{fig_carotid}), so the selection matrix for force is \(^{ee}\bm S_f = diag([0,0,1,0,0,0])\),
and the remaining dimensions are controlled with impedance; the selection matrix for impedance is \(^{ee}\bm S_i = diag([1,1,0,1,1,1])\).
After transformation, the selection matrix in base frame used in the main task controller (\ref{eq_tau1_regu}) \(\bm S_i, \bm S_f\) is obtained.

Note that the inference frequency of the upper navigation level is nearly 10 Hz, and the control frequency is 1kHz, so we adopted a low-pass filter to make the actions smoother.
\begin{table}
\centering
\caption{Tracking and Force Stats on Volunteers}
\vspace{-3mm}
\label{tab:volunteers_stats}
\begin{tblr}{
  width = \linewidth,
  colspec = {Q[165]Q[104]Q[179]Q[160]Q[213]Q[100]},
  column{4} = {c},
  column{5} = {c},
  column{6} = {c},
  cell{1}{4} = {c=3}{0.473\linewidth},
  cell{2}{3} = {c},
  cell{3}{3} = {c},
  cell{4}{3} = {c},
  cell{5}{3} = {c},
  cell{6}{3} = {c},
  hline{1,3,7} = {-}{},
  hline{2} = {2-6}{},
}
        &      & {Tracking\\~Error (m)} & Wrench    &             &       \\
        &      & Transverse             & Force (N) & Torque (Nm) & $dF_z/dt$ (N/s) \\
Unknown & Mean & 0.0135                 & 2.503     & 0.259       & 0.159 \\
        & Max  & 0.0221                 & 3.699     & 0.526       &       \\
Known   & Mean & 0.0104                 & 1.889     & 0.262       & 0.154 \\
        & Max  & 0.0178                 & 3.728     & 0.533       &       
\end{tblr}
\vspace{-4mm}
\end{table}
\begin{figure*}[!ht]
  \centering
  \subcaptionbox{}
    {\includegraphics[width=0.24\linewidth]{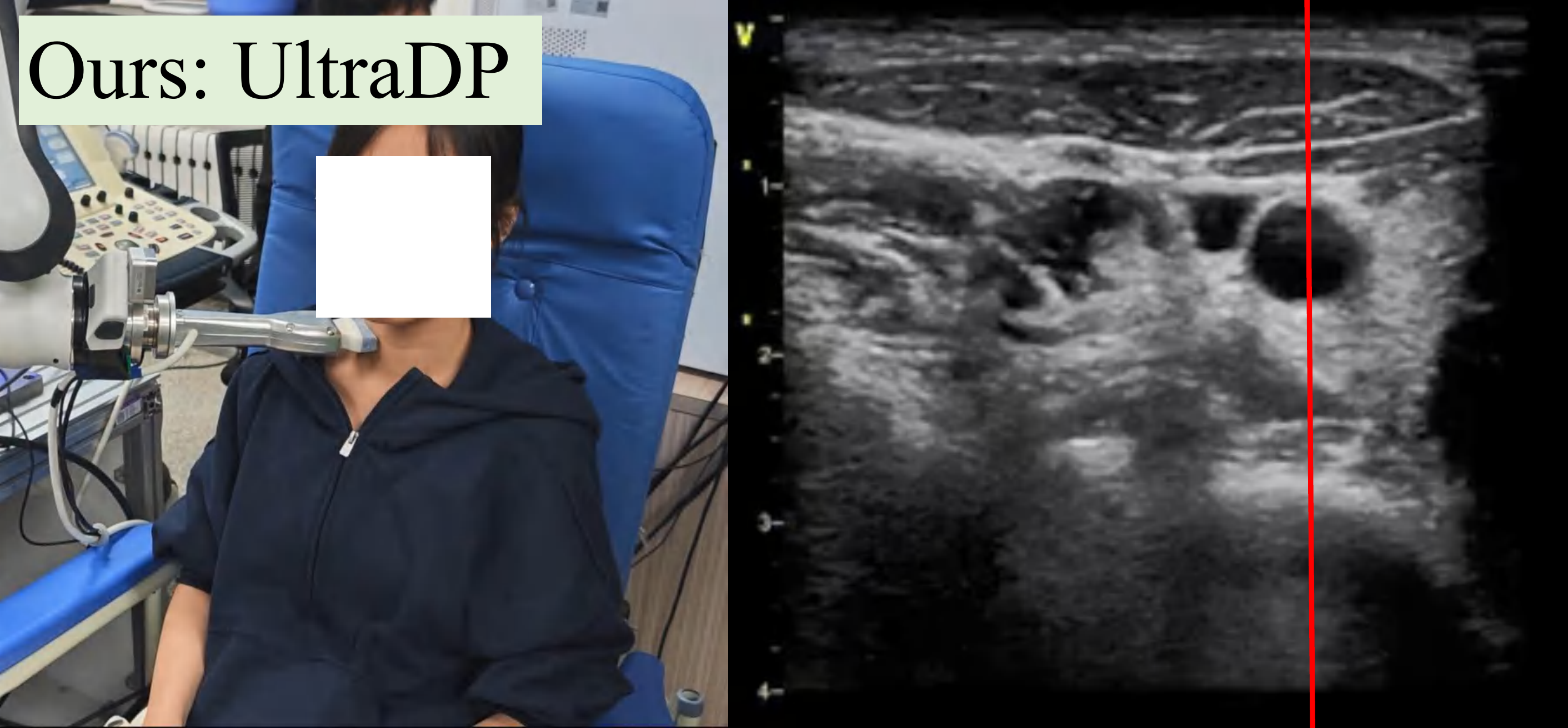}}
  \subcaptionbox{}
    {\includegraphics[width=0.24\linewidth]{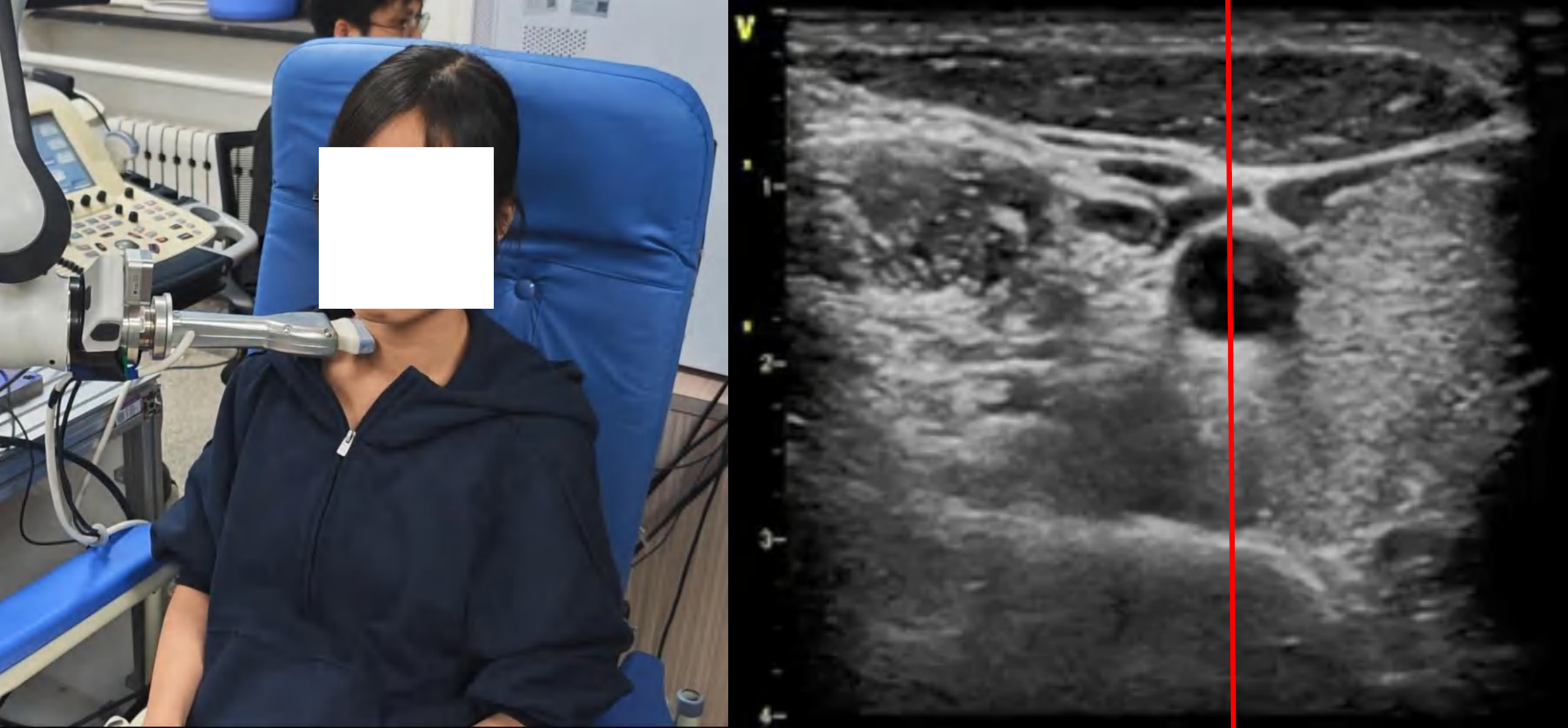}}
  \subcaptionbox{}
    {\includegraphics[width=0.24\linewidth]{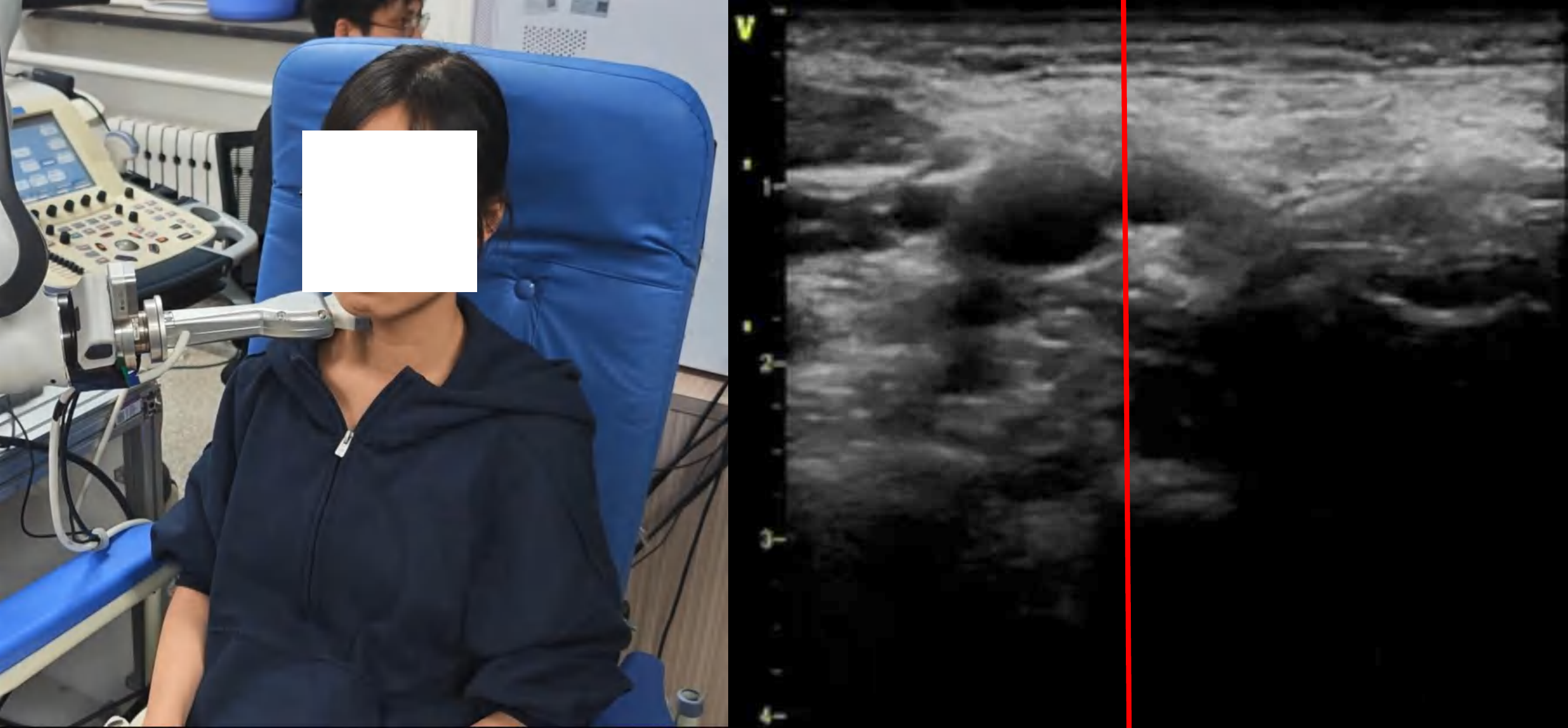}}
  \subcaptionbox{}
    {\includegraphics[width=0.24\linewidth]{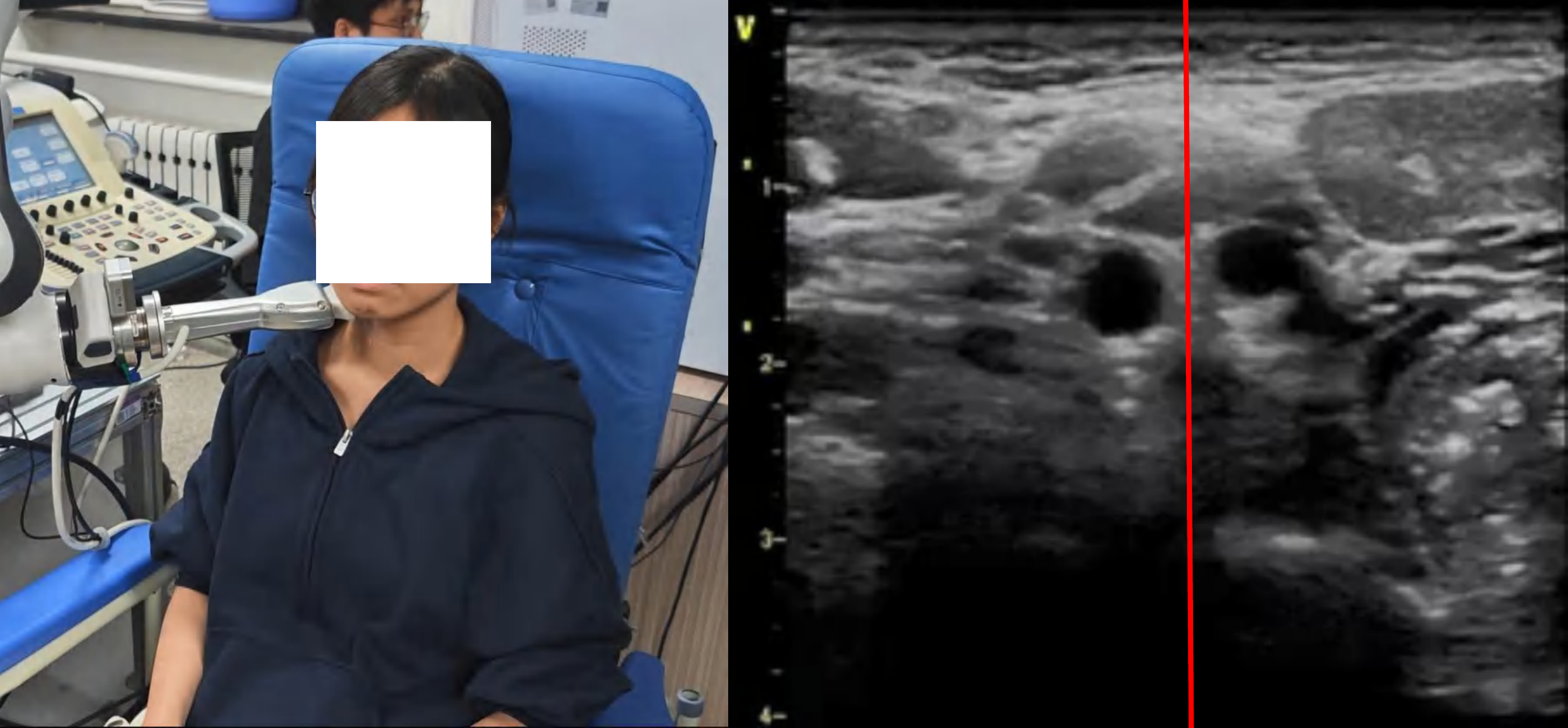}}
  \subcaptionbox{}
    {\includegraphics[width=0.24\linewidth]{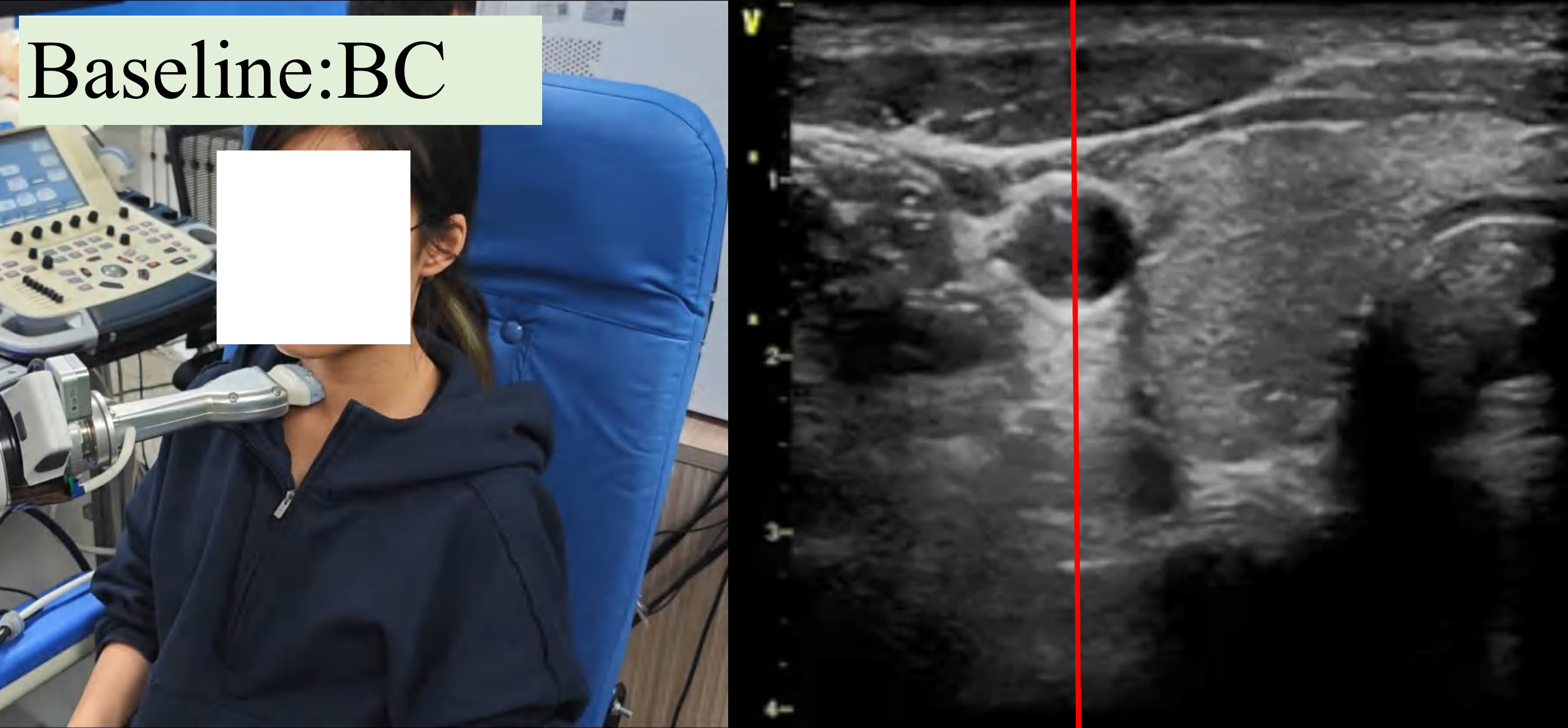}}
  \subcaptionbox{}
    {\includegraphics[width=0.24\linewidth]{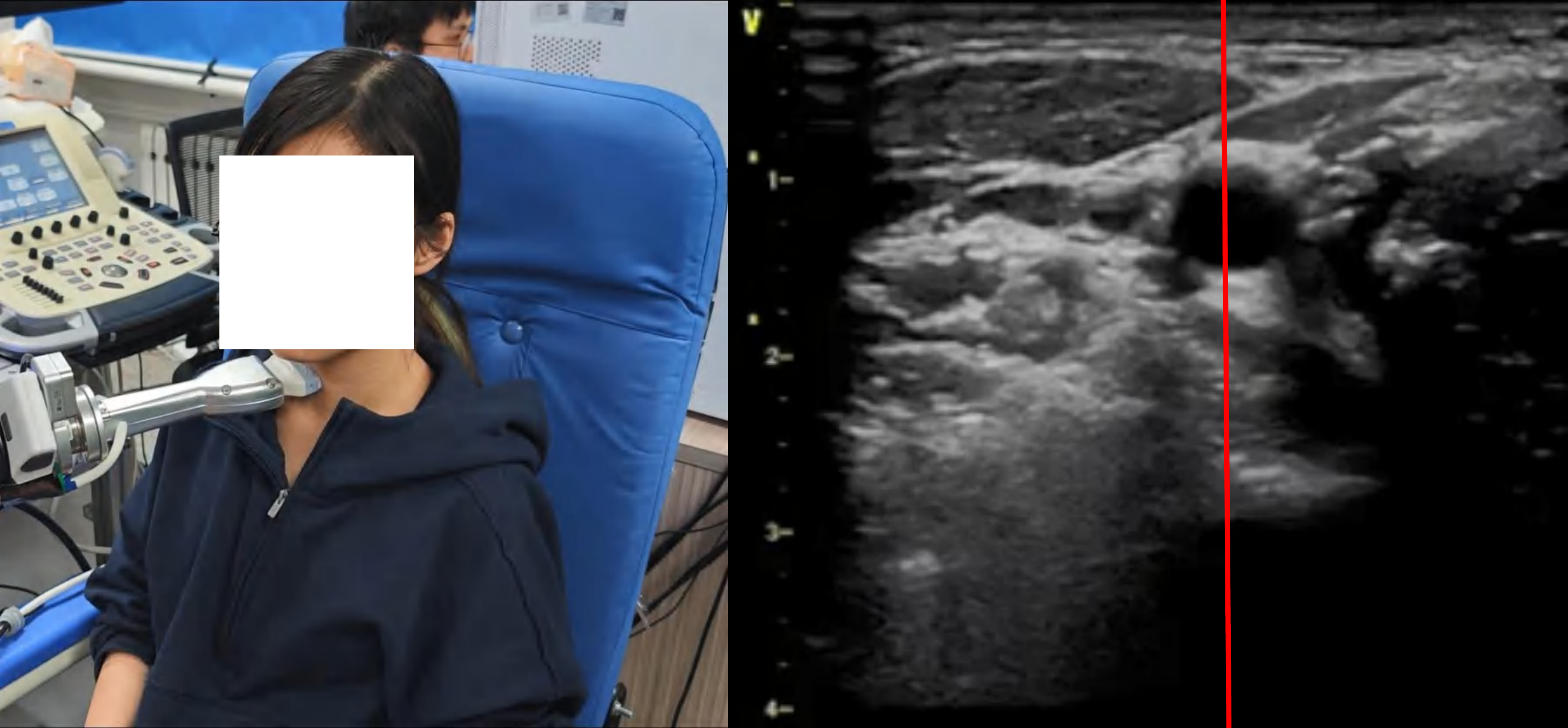}}
  \subcaptionbox{}
    {\includegraphics[width=0.24\linewidth]{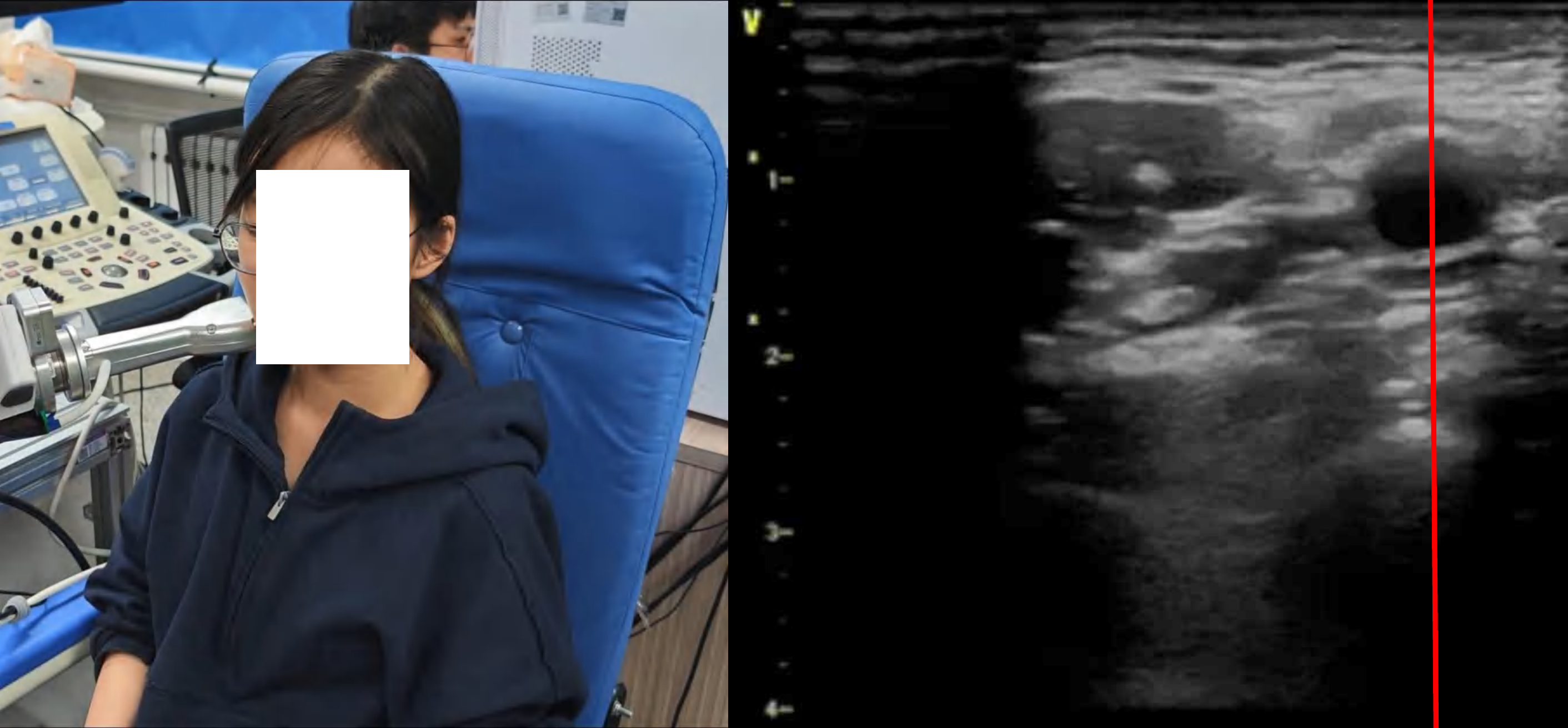}}
  \subcaptionbox{}
    {\includegraphics[width=0.24\linewidth]{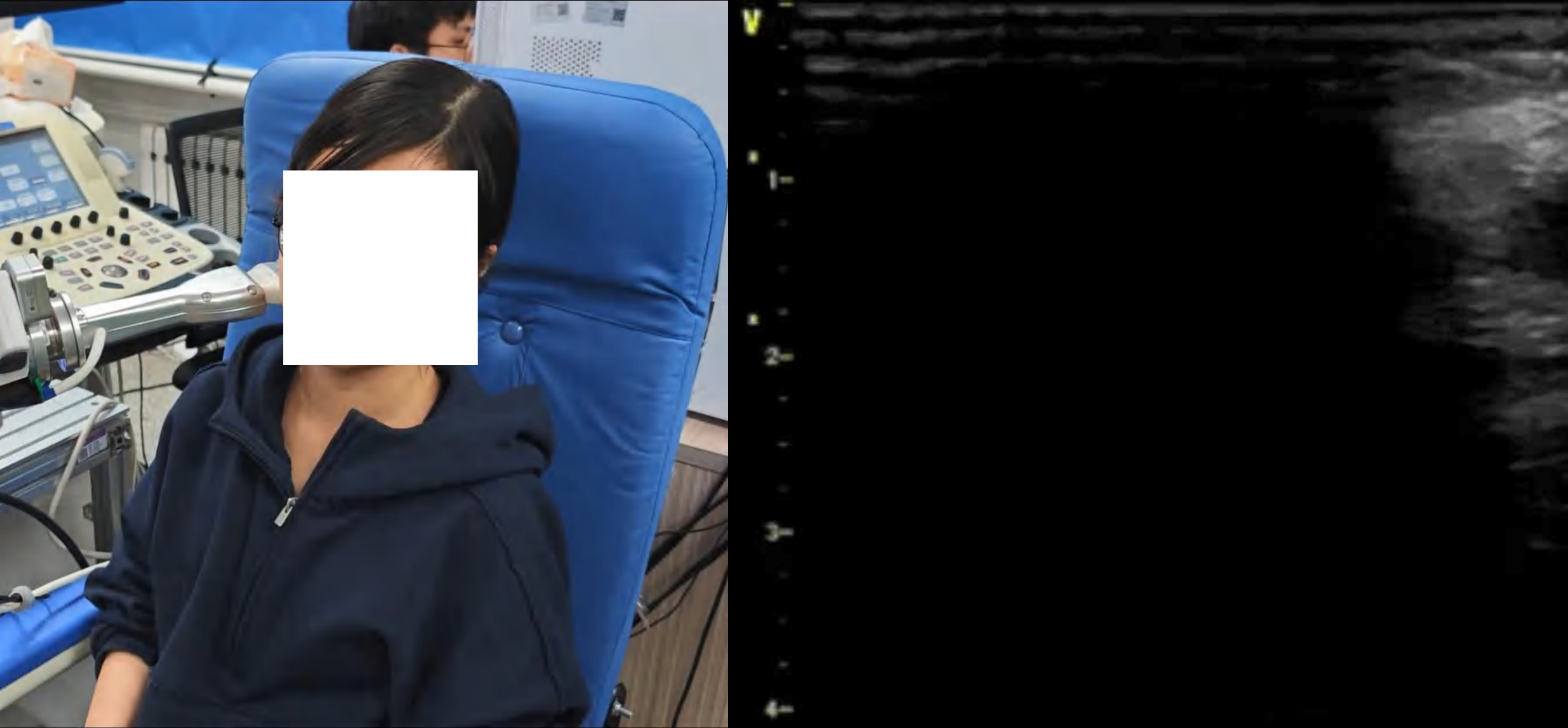}}
  \subcaptionbox{}
    {\includegraphics[width=0.24\linewidth]{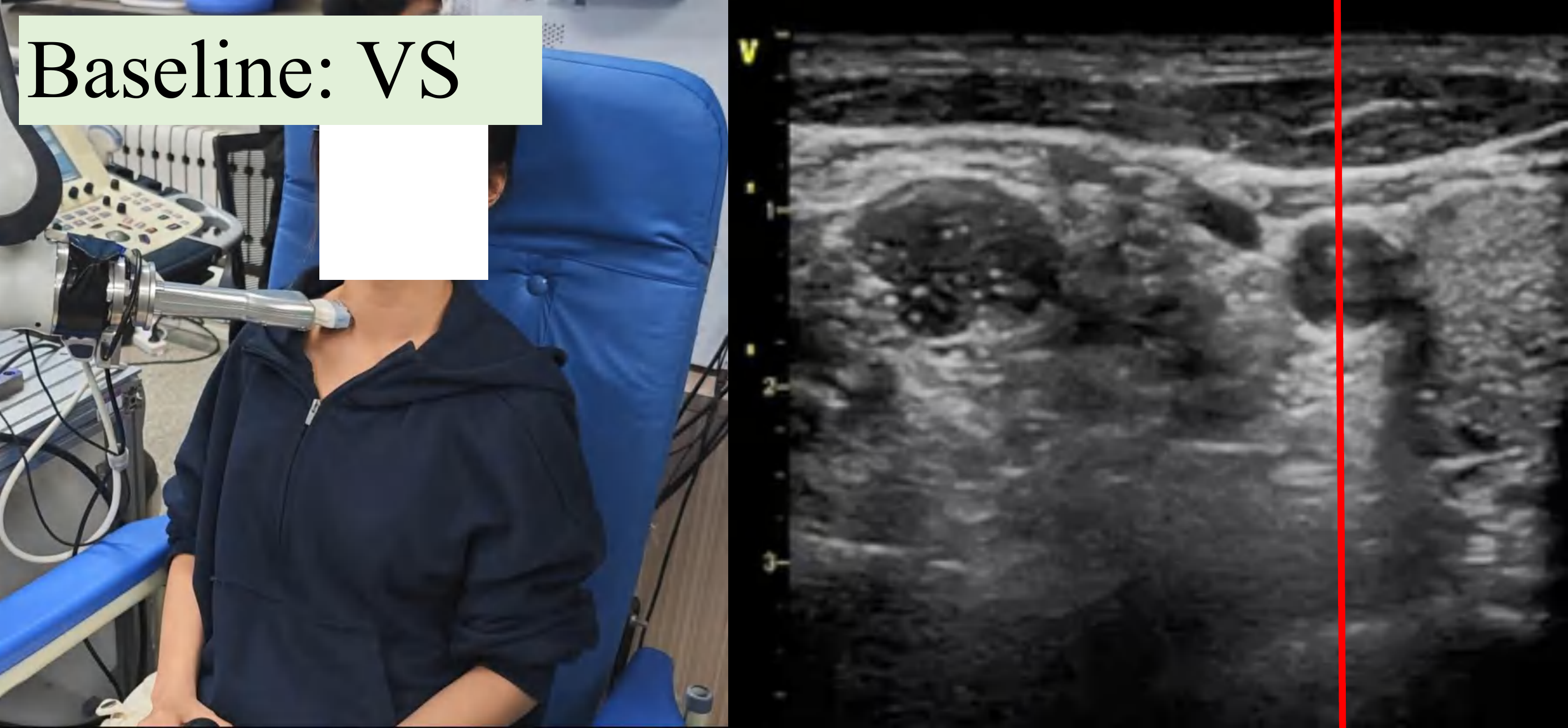}}
  \subcaptionbox{}
    {\includegraphics[width=0.24\linewidth]{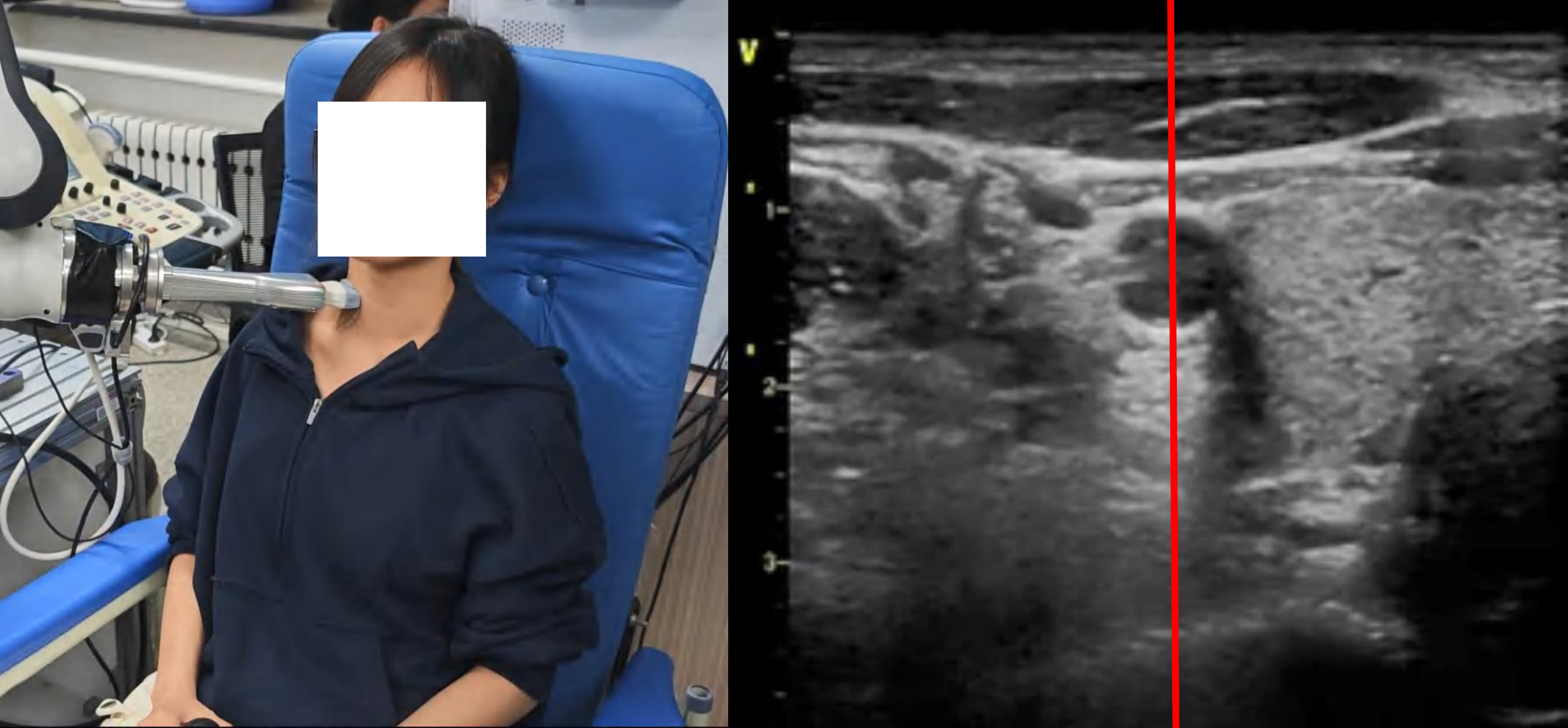}}
  \subcaptionbox{}
    {\includegraphics[width=0.24\linewidth]{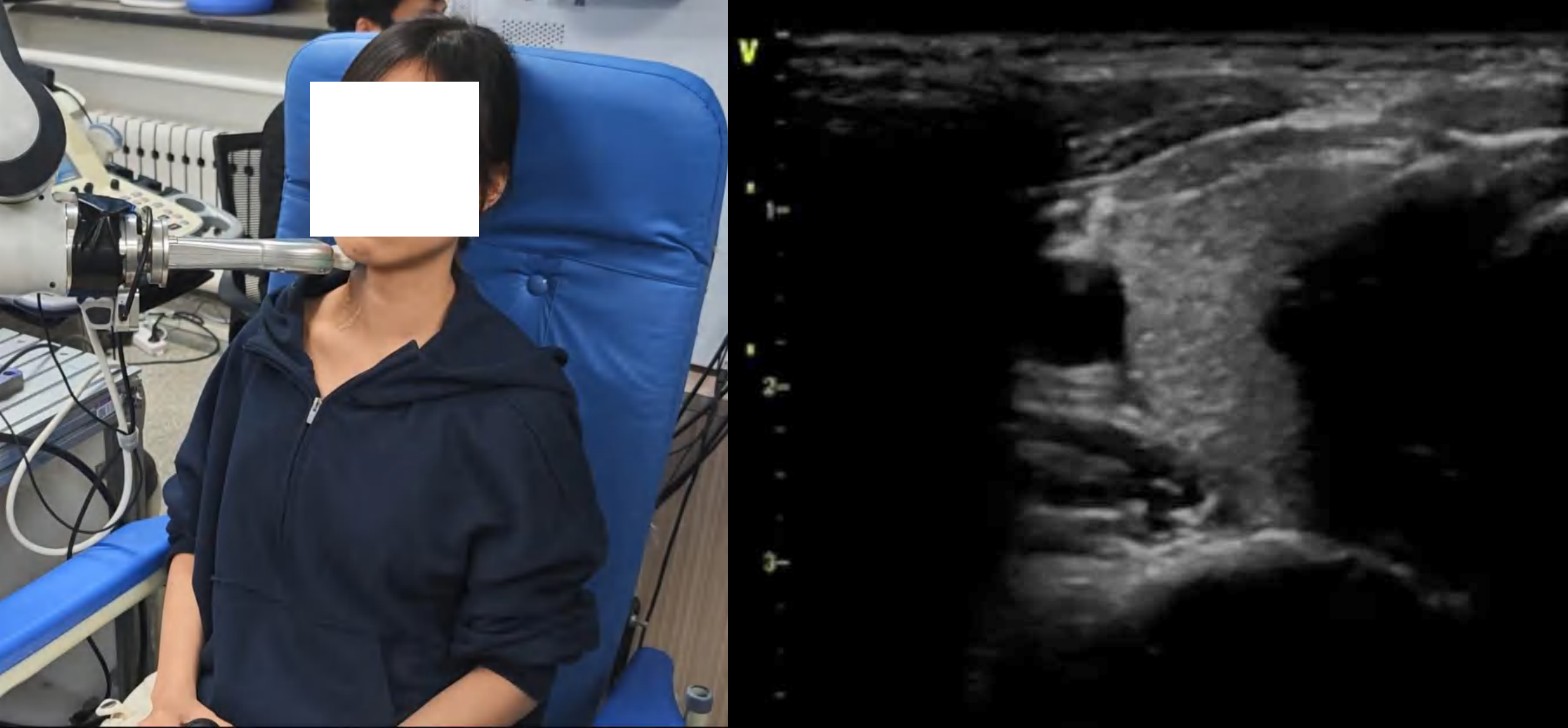}}
  \subcaptionbox{}
    {\includegraphics[width=0.24\linewidth]{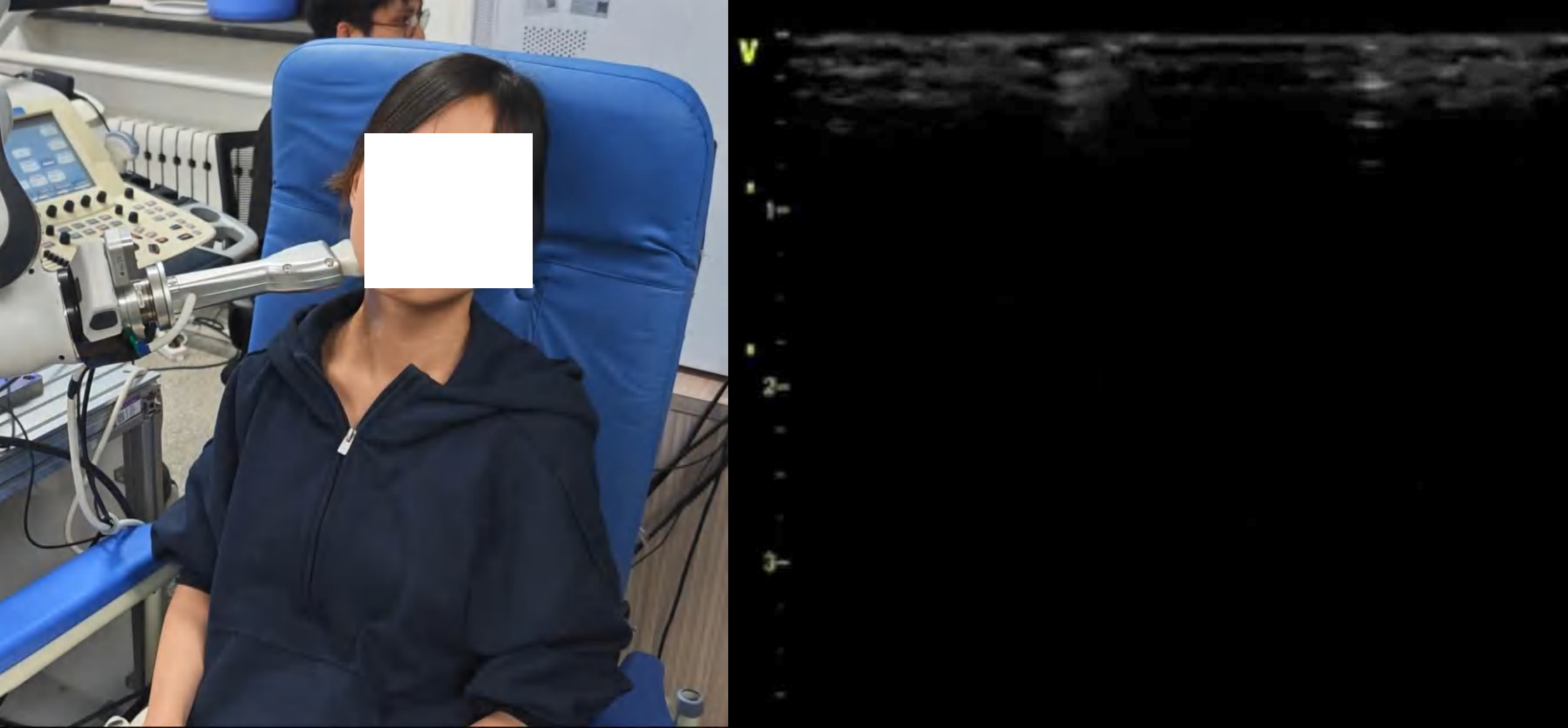}}
\caption{Real-World Experiment - Snapshots and ultrasound images, the red lines are the output of our pretrained regressor. No red line means the network can not detect the artery position.
First row: UltraDP.
(a) The policy began, and the artery was on the right of the image; (b) The policy output actions to guide the robot to make the artery center while going upwards; (c) The policy detected the bifurcation of the artery; (d) The external and internal arteries were clear in the image, and the scanning was over.
Second row: Baseline, behavior cloning. 
(e)-(h): The BC policy (baseline 1) did not center the artery; And in the end, the policy drove the probe away from the neck, showing the unsatisfying generalization ability.
Third row: Baseline, visual serving.
(i)-(l) The VS controller (baseline 2) had the ability to center the artery; however, because some parameters like \textit{offset\_z} did not suit the female, the probe detached her neck, and the regressor couldn't work when the image was incomplete; at last the probe kept going up, losing the image and hit her in the jaw.
}
\label{fig_exp2_snapshots}
\vspace{-3mm}
\end{figure*}

\section{Experiments}
Our hardware system consists of five parts, as shown in Fig. \ref{fig_exp_setup}: a Franka manipulator, an ultrasound machine with a probe, an ATI mini 40 force/torque sensor connected between the arm flange and the ultrasound probe, a RealSense D405 camera mounted on the arm flange, a control computer with an AMD Ryzen 5 5600G CPU and a Nvidia RTX 3060 GPU in Ubuntu 20.04 system. The navigation module operates at 10 Hz, while the hybrid force-impedance controller runs at 1kHz. More experiment details can be found on our website\footnote{\url{https://ultradp.github.io/}}.

We conducted a series of simulation studies and experiments to validate: (\romannumeral1) UltraDP’s learned understanding of scanning knowledge in simulation (in Sec. \ref{sec_exp1_sim}); (\romannumeral2) the generalization ability of UltraDP system on unseen human subjects compared to a rule-based Visual Servoing (VS) method and a classic learning-based Behavior Cloning (BC) method in real-world (in Sec. \ref{sec_exp2_comparative}); (\romannumeral3) the effect of wrench and multi-sensory observation (in Sec. \ref{sec_exp3_ablation}); (\romannumeral4) subjects' feelings on UltraDP system based on their scanning experience (in Sec. \ref{sec_exp4_subjective}).

\subsection{Simulation Studies}\label{sec_exp1_sim}
We first evaluate our policy on a set of trajectories randomly sampled from the training dataset in simulation. Then, we asked the same sonographer to demonstrate several new volunteers (including both male and female) that were completely unseen to our policy. The new demonstrations were then used in evaluation to assess the model's generalization ability: the trajectories were converted to observations and input to the policy, then the action of next time predicted by the policy was taken to be compared with the real trajectory. The predicted force and torque are shown in Tab. \ref{tab:volunteers_stats}, where the metric \(dF_z/dt\) denotes the average changing rate of predicted force (also referred to as the force rate) throughout the entire trajectory, serving as an indicator of the subject's comfort.

As shown in Table \ref{tab:volunteers_stats}, we can tell that compared to ``known'' trajectories, the performance of ``unknown'' did decline, but not much, indicating UltraDP had a certain degree of generalization ability.

The force prediction on unknown subjects increased within a small range, which can be interpreted as an attempt to acquire a better image when unfamiliar scenarios occur. The desired force predicted by UltraDP is within the range of 4N and maintains minimal fluctuation in all trajectories, showcasing steady performance and generalization capability.

\subsection{Real-World Experiment}\label{sec_exp2_comparative}

\begin{figure}[!tb]
  \centering
  \includegraphics[width=1.0\linewidth]{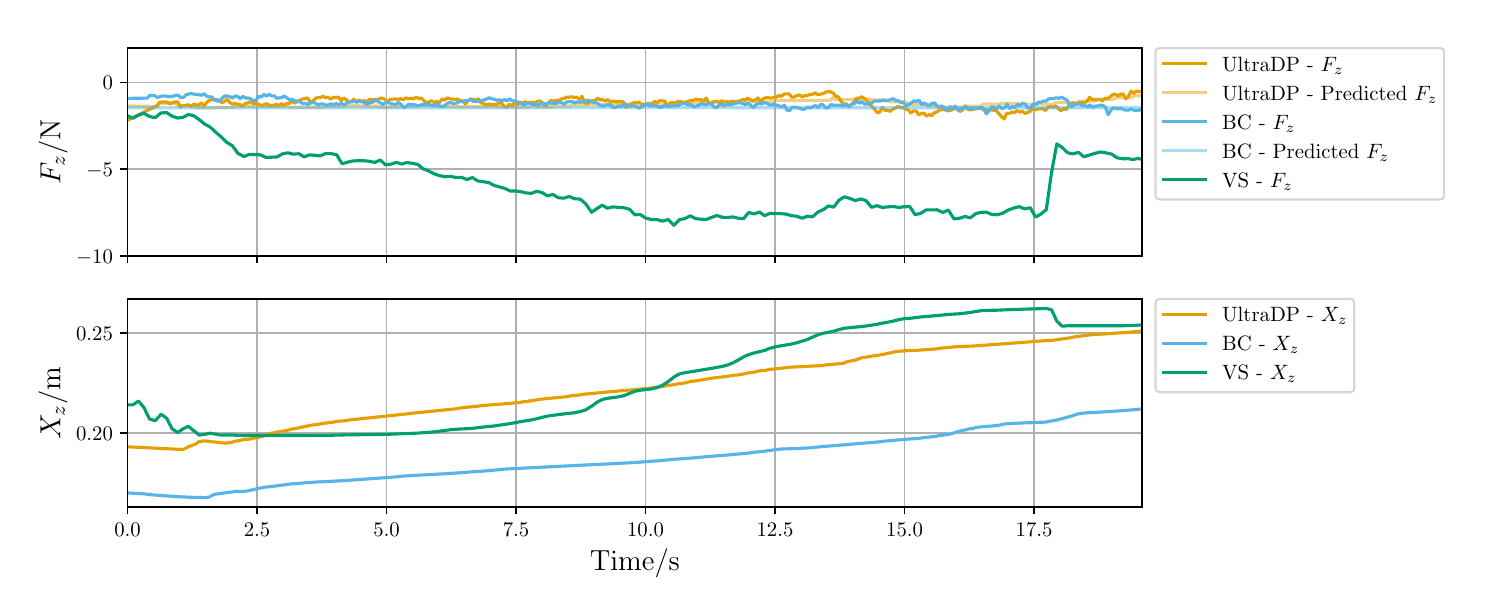}
\caption{Real-World Experiment -  $F_z$ in probe frame and $X_z$ in base frame during the whole process: When the experiment started, the VS method exerted excessive force on the volunteer's clavicle, causing a large contact force during this stage. The VS pushed the neck and the jaw of the volunteer, reflecting in the increasing force. The BC method failed to keep track of the target, moving sideway during the scanning. In contrast, our method guided the probe to maintain a steady ultrasound image while effectively preventing excessive contact with the patient.}
\label{fig_exp2_}
\end{figure}

\begin{table*}
\centering
\caption{Comparative Experiments with Baseline on 20 Real-World Ultrasound Scanning Trials}
\vspace{-3mm}
\label{tab_exp2_results}
\begin{tblr}{
  cells = {c},
  hline{1-2,5} = {-}{},
}
                                                  & Success Rate ($\uparrow$) & {Mean Landmark Distance\\to Center on $x$ axis\\during Transverse($\downarrow$)/pixel} & {Mean SSIM Distance\\ with Human Expert\\Videos ($\uparrow$)} & {Mean Expert\\Score($\uparrow$)} & {Mean Contact Force\\Rate on $z$ axis\\~($dF_z/dt$))($\downarrow$)/(N/s)} \\
\textbf{UltraDP}                                  & \textbf{19/20}             & \textbf{5.71}                                                                          & \textbf{76.83\%}                                              & \textbf{7.11}                    & \textbf{0.187}                                                            \\
Baseline VS\cite{yan2024unified} & 17/20                      & 6.52                                                                                   & 71.24\%                                                       & 6.76                             & 0.329                                                                     \\
Baseline BC                                       & 6/20                      & 25.44                                                                                  & 62.19\%                                                       & 6.14                             & 0.189                                                                     
\end{tblr}
\vspace{-3mm}
\end{table*}

We evaluated UltraDP alongside a rule-based Visual Servo (VS) method modified from \cite{yan2024unified} and Behavior Cloning (BC) as baselines on two female and two male volunteers not included in the training data, conducting five trials per participant. 
Note that \cite{yan2024unified} only proposed a cartesian space impedance controller. We added an ultrasound image space servoing term for fairness, with the aid of our pretrained network: If an artery was detected, the servoing term was activated, otherwise not.
The model parameters, such as artery angle, target offsets, and neck length, were calibrated on testers and then averaged. The baseline BC was trained with the same data and implemented with the same hybrid force impedance controller.

Statistical results are shown in Tab. \ref{tab_exp2_results}. We define "Success Rate" as whether the ultrasound image shows a clear centered transverse view more than 60\% of the scanning period and the probe stops upon detecting a bifurcation.
The metric "Landmark Distance to Center on $x$ axis" is calculated with the help of our pretrained encoder. ``SSIM Distance with Human Expert Videos'' measures the Structural Similarity Index (SSIM) between experiment videos and an expert video recorded by a sonographer scanning the same subject after aligning the video timelines. Finally, the "Expert Score" is obtained by sending the videos single-blindly to expert sonographers, who rated them from 0 to 10.
Across these five metrics, UltraDP consistently outperformed the rule-based method and BC, demonstrating superior generalization across different subjects.

Snapshots of one typical trial and the force and pose information are presented in Fig. \ref{fig_exp2_snapshots}, \ref{fig_exp2_}, involving a female volunteer unknown to the policy.  The $F_z$ curve shows the significant force exerted by VS method as it pushes on the volunteer’s clavicle and neck in the beginning. While the BC method with our controller maintained a consistent force, it failed to track the landmark, producing incorrect movements that drove the probe away from the volunteer’s neck. In contrast, UltraDP maintained stable contact and accurately predicted a suitable force for scanning.


\subsection{Ablation Study}\label{sec_exp3_ablation}
We did an ablation study of the modalities of the observation, the results are listed in Tab. \ref{tab_exp3_results}. ``UltraDP obs w/o \{pose, wrench\}'' showed no convergence during training, so we didn't test it on a human subject. We could conclude that both pose and wrench information were important for success, but wrench information appeared to be even more critical in UltraDP's framework.

\begin{figure}[!t]
    \centering
    \includegraphics[width=1.02\linewidth]{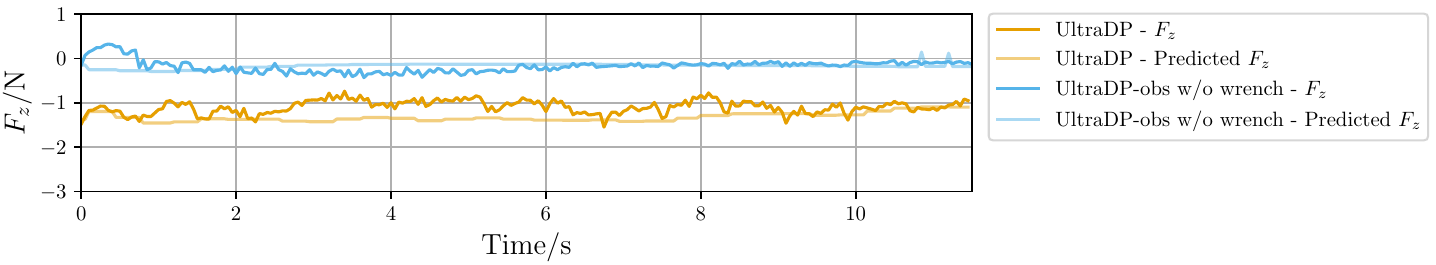}
    \caption{Ablation Study - The force and torque of UltraDP and UltraDP obs w/o force. UltraDP successfully executed the desired trajectory with stable contact behavior, while the model trained without force input failed to estimate the correct wrench, leading to the probe losing contact with the neck.}
    \label{exp3_fig}
\end{figure}

During the testing, we also found that the failure modes of the same method were often similar. For example, ``obs w/o wrench'' frequently failed to make contact with the neck because it outputs an inappropriate predicted contact force to the low-level controller, as illustrated in Fig. \ref{exp3_fig}. On the other hand, ``obs w/o pose'' usually managed to make contact but struggled to keep the artery centered in the ultrasound image, often causing the landmark to drift out of view and leading to failure. This showed that wrench information ensured proper contact, while pose information helped keep the artery centered in our framework.
\begin{table}
\centering
\caption{Ablation Study of Components in Observation}
\label{tab_exp3_results}
\vspace{-2mm}
\begin{tblr}{
  cells = {c},
  hline{1-2,6} = {-}{},
}
                                  & Success Rate ($\uparrow$) \\
\textbf{UltraDP}                  & \textbf{19/20}            \\
UltraDP, obs w/o wrench           & 0/20                      \\
UltraDP, obs w/o pose             & 8/20                      \\
UltraDP, obs w/o \{pose, wrench\} & /                         
\end{tblr}
\vspace{-3mm}
\end{table}

\subsection{Subjective Study}\label{sec_exp4_subjective}
\begin{figure}[!tb]
    \centering
    \includegraphics[width=0.95\linewidth]{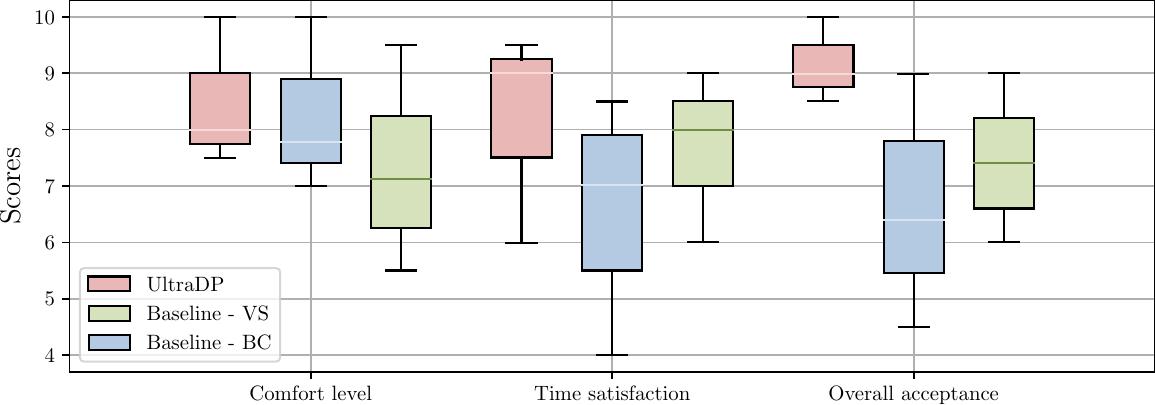}
    \caption{Subjective Study - Statistic results from questionnaires}
    \label{fig_exp4}
\end{figure}

To assess the volunteers' evaluation on our UltraDP system, we did a questionnaire survey after they experienced the robotic scanning using UltraDP and two baseline methods. A total of 24 valid responses were collected, with the results presented in Fig. \ref{fig_exp4}. The feedback indicates that UltraDP excels in comfort and efficiency, making it well-suited for human-in-the-loop scenarios.

\section{Conclusion}
This paper deals with the generalization problem in autonomous robotic ultrasound scanning, which is important as it guarantees the practical value of the robot across multiple human subjects with different backgrounds and medical conditions. To improve the generalization ability, we refer to the recent progress of diffusion policy and develop UltraDP, a force-aware diffusion policy that can better represent multi-modal action distributions, utilize the multi-sensory information and generate smooth scanning trajectories. Such a trajectory is further tracked and hence realized with a hybrid force impedance control scheme so that the safety of physical interaction is achieved. The safety, effectiveness, and generalization of the developed robot are validated in real-world experiments. The scaling-up validation will be our future work.

\bibliographystyle{ieeetr}
\bibliography{main}

@article{jiang-2025,
	author = {Jiang, Haojun and Zhao, Andrew and Yang, Qian and Yan, Xiangjie and Wang, Teng and Wang, Yulin and Jia, Ning and Wang, Jiangshan and Wu, Guokun and Yue, Yang and Luo, Shaqi and Wang, Huanqian and Ren, Ling and Chen, Siming and Liu, Pan and Yao, Guocai and Yang, Wenming and Song, Shiji and Li, Xiang and He, Kunlun and Huang, Gao},
	journal = {Nature Communications},
	month = {8},
	number = {1},
	pages = {7893},
	title = {{Towards expert-level autonomous carotid ultrasonography with large-scale learning-based robotic system}},
	volume = {16},
	year = {2025},
	doi = {10.1038/s41467-025-62865-w},
	url = {https://doi.org/10.1038/s41467-025-62865-w},
}

@INPROCEEDINGS{10802113,
  author={Ranne, Alex and Kuang, Liming and Velikova, Yordanka and Navab, Nassir and Baena, Ferdinando Rodriguez Y},
  booktitle={2024 IEEE/RSJ International Conference on Intelligent Robots and Systems (IROS)}, 
  title={CathFlow: Self-Supervised Segmentation of Catheters in Interventional Ultrasound Using Optical Flow and Transformers}, 
  year={2024},
  volume={},
  number={},
  pages={2436-2443},
  doi={10.1109/IROS58592.2024.10802113}}

@article{huang2021towards,
  title={Towards fully autonomous ultrasound scanning robot with imitation learning based on clinical protocols},
  author={Huang, Yanwei and Xiao, Wei and Wang, Chuyang and Liu, Hengli and Huang, Rui and Sun, Zhenglong},
  journal={IEEE Robotics and Automation Letters},
  volume={6},
  number={2},
  pages={3671--3678},
  year={2021},
  publisher={IEEE}
}

@inproceedings{si2023unified,
  title={A unified deep imitation learning and control framework for robot-assisted sonography},
  author={Si, Weiyong and Guo, Cheng and Wang, Ning and Yang, Minyu and Harris, Rebecca and Yang, Chenguang},
  booktitle={2023 IEEE International Conference on Development and Learning (ICDL)},
  pages={318--323},
  year={2023},
  organization={IEEE}
}

@inproceedings{deng2021learning,
  title={Learning robotic ultrasound scanning skills via human demonstrations and guided explorations},
  author={Deng, Xutian and Chen, Yiting and Chen, Fei and Li, Miao},
  booktitle={2021 IEEE International Conference on Robotics and Biomimetics (ROBIO)},
  pages={372--378},
  year={2021},
  organization={IEEE}
}

@article{chi2023diffusion,
  title={Diffusion policy: Visuomotor policy learning via action diffusion},
  author={Chi, Cheng and Xu, Zhenjia and Feng, Siyuan and Cousineau, Eric and Du, Yilun and Burchfiel, Benjamin and Tedrake, Russ and Song, Shuran},
  journal={The International Journal of Robotics Research},
  pages={02783649241273668},
  year={2023},
  publisher={SAGE Publications Sage UK: London, England}
}

@inproceedings{jiang2024cardiac,
  title = {Cardiac {{Copilot}}: {{Automatic Probe Guidance}} for~{{Echocardiography}} with~{{World Model}}},
  booktitle = {Medical {{Image Computing}} and {{Computer Assisted Intervention}} -- {{MICCAI}} 2024},
  author = {Jiang, Haojun and Sun, Zhenguo and Jia, Ning and Li, Meng and Sun, Yu and Luo, Shaqi and Song, Shiji and Huang, Gao},
  editor = {Linguraru, Marius George and Dou, Qi and Feragen, Aasa and Giannarou, Stamatia and Glocker, Ben and Lekadir, Karim and Schnabel, Julia A.},
  year = {2024},
  pages = {190--199},
  publisher = {Springer Nature Switzerland},
  address = {Cham},
  doi = {10.1007/978-3-031-72378-0_18},
  isbn = {978-3-031-72378-0},
  langid = {english}
}

@inproceedings{jiang2025structureaware,
  title = {Structure-Aware {{World Model}} for {{Probe Guidance}} via {{Large-scale Self-supervised Pre-train}}},
  booktitle = {Simplifying {{Medical Ultrasound}}},
  author = {Jiang, Haojun and Li, Meng and Sun, Zhenguo and Jia, Ning and Sun, Yu and Luo, Shaqi and Song, Shiji and Huang, Gao},
  editor = {Gomez, Alberto and Khanal, Bishesh and King, Andrew and Namburete, Ana},
  year = {2025},
  pages = {58--67},
  publisher = {Springer Nature Switzerland},
  address = {Cham},
  doi = {10.1007/978-3-031-73647-6_6},
  isbn = {978-3-031-73647-6},
  langid = {english}
}

@misc{jiang2024sequenceaware,
  title = {Sequence-Aware {{Pre-training}} for {{Echocardiography Probe Guidance}}},
  author = {Jiang, Haojun and Sun, Zhenguo and Sun, Yu and Jia, Ning and Li, Meng and Luo, Shaqi and Song, Shiji and Huang, Gao},
  year = {2024},
  number = {arXiv:2408.15026},
  eprint = {2408.15026},
  primaryclass = {cs},
  publisher = {arXiv},
  doi = {10.48550/arXiv.2408.15026},
  urldate = {2025-02-23},
  archiveprefix = {arXiv}
}

@ARTICLE{10684288,
  author={Wang, Ziwen and Han, Yingying and Zhao, Baoliang and Xie, Haiqin and Yao, Liang and Li, Bing and Meng, Max Q.-H. and Hu, Ying},
  journal={IEEE Transactions on Medical Robotics and Bionics}, 
  title={Autonomous Robotic System for Carotid Artery Ultrasound Scanning With Visual Servo Navigation}, 
  year={2024},
  volume={6},
  number={4},
  pages={1436-1447},
  doi={10.1109/TMRB.2024.3464109}}

@inproceedings{yan2024unified,
  title={A Unified Interaction Control Framework for Safe Robotic Ultrasound Scanning with Human-Intention-Aware Compliance},
  author={Yan, Xiangjie and Luo, Shaqi and Jiang, Yongpeng and Yu, Mingrui and Chen, Chen and Zhu, Senqiang and Huang, Gao and Song, Shiji and Li, Xiang},
  booktitle={2024 IEEE/RSJ International Conference on Intelligent Robots and Systems (IROS)},
  pages={14004--14011},
  year={2024},
  organization={IEEE}
}

@article{huang2018fully,
  title={Fully automatic three-dimensional ultrasound imaging based on conventional B-scan},
  author={Huang, Qinghua and Wu, Bowen and Lan, Jiulong and Li, Xuelong},
  journal={IEEE transactions on biomedical circuits and systems},
  volume={12},
  number={2},
  pages={426--436},
  year={2018},
  publisher={IEEE}
}

@article{yan2023multi,
  title={Multi-Modal Interaction Control of Ultrasound Scanning Robots with Safe Human Guidance and Contact Recovery},
  author={Yan, Xiangjie and Jiang, Yongpeng and Wu, Guokun and Chen, Chen and Huang, Gao and Li, Xiang},
  journal={arXiv preprint arXiv:2302.05685},
  year={2023}
}

@article{harrison2015work,
  title={Work-related musculoskeletal disorders in ultrasound: Can you reduce risk?},
  author={Harrison, Gill and Harris, Allison},
  journal={Ultrasound},
  volume={23},
  number={4},
  pages={224--230},
  year={2015},
  publisher={SAGE Publications Sage UK: London, England}
}

@inproceedings{goel2022autonomous,
  title={Autonomous ultrasound scanning using bayesian optimization and hybrid force control},
  author={Goel, Raghavv and Abhimanyu, Fnu and Patel, Kirtan and Galeotti, John and Choset, Howie},
  booktitle={2022 International Conference on Robotics and Automation (ICRA)},
  pages={8396--8402},
  year={2022},
  organization={IEEE}
}

@misc{wu2024tacdiffusion,
  title = {{{TacDiffusion}}: {{Force-domain Diffusion Policy}} for {{Precise Tactile Manipulation}}},
  author = {Wu, Yansong and Chen, Zongxie and Wu, Fan and Chen, Lingyun and Zhang, Liding and Bing, Zhenshan and Swikir, Abdalla and Knoll, Alois and Haddadin, Sami},
  year = {2024},
  number = {arXiv:2409.11047},
  eprint = {2409.11047},
  primaryclass = {cs},
  publisher = {arXiv},
  doi = {10.48550/arXiv.2409.11047},
  urldate = {2025-02-23},
  archiveprefix = {arXiv}
}

@article{ze20243d,
  title={3d diffusion policy: Generalizable visuomotor policy learning via simple 3d representations},
  author={Ze, Yanjie and Zhang, Gu and Zhang, Kangning and Hu, Chenyuan and Wang, Muhan and Xu, Huazhe},
  journal={arXiv preprint arXiv:2403.03954},
  year={2024}
}

@ARTICLE{10878457,
  author={Bi, Yuan and Su, Yang and Navab, Nassir and Jiang, Zhongliang},
  journal={IEEE Robotics and Automation Letters}, 
  title={Gaze-Guided Robotic Vascular Ultrasound Leveraging Human Intention Estimation}, 
  year={2025},
  volume={10},
  number={4},
  pages={3078-3085},
  doi={10.1109/LRA.2025.3539546}}

@INPROCEEDINGS{10715958,
  author={Wu, Guokun and Luo, Shaqi and Huang, Gao and Li, Xiang},
  booktitle={2024 International Conference on Advanced Robotics and Mechatronics (ICARM)}, 
  title={Vision-Based Detection and Real-Time Adaptive Control Schemes for Autonomous Ultrasound Scanning Robots}, 
  year={2024},
  volume={},
  number={},
  pages={631-636},
  doi={10.1109/ICARM62033.2024.10715958}}

@article{hou2024adaptive,
  title={Adaptive Compliance Policy: Learning Approximate Compliance for Diffusion Guided Control},
  author={Hou, Yifan and Liu, Zeyi and Chi, Cheng and Cousineau, Eric and Kuppuswamy, Naveen and Feng, Siyuan and Burchfiel, Benjamin and Song, Shuran},
  journal={arXiv preprint arXiv:2410.09309},
  year={2024}
}

@article{liu2024forcemimic,
  title={ForceMimic: Force-Centric Imitation Learning with Force-Motion Capture System for Contact-Rich Manipulation},
  author={Liu, Wenhai and Wang, Junbo and Wang, Yiming and Wang, Weiming and Lu, Cewu},
  journal={arXiv preprint arXiv:2410.07554},
  year={2024}
}

@misc{mayo,
  title = {Mayo Clinic: Carotid Ultrasound},
  howpublished = {https://www.mayoclinic.org/tests-procedures/carotid-ultrasound/about/pac-20393399},
  note = {Accessed: 2025-02-22},
}

@article{duan2022ultrasound,
  title={Ultrasound-guided assistive robots for scoliosis assessment with optimization-based control and variable impedance},
  author={Duan, Anqing and Victorova, Maria and Zhao, Jingyuan and Sun, Yuxiang and Zheng, Yongping and Navarro-Alarcon, David},
  journal={IEEE Robotics and Automation Letters},
  volume={7},
  number={3},
  pages={8106--8113},
  year={2022},
  publisher={IEEE}
}

@article{ho2020denoising,
  title={Denoising diffusion probabilistic models},
  author={Ho, Jonathan and Jain, Ajay and Abbeel, Pieter},
  journal={Advances in neural information processing systems},
  volume={33},
  pages={6840--6851},
  year={2020}
}

@inproceedings{zhou2019continuity,
  title={On the continuity of rotation representations in neural networks},
  author={Zhou, Yi and Barnes, Connelly and Lu, Jingwan and Yang, Jimei and Li, Hao},
  booktitle={Proceedings of the IEEE/CVF conference on computer vision and pattern recognition},
  pages={5745--5753},
  year={2019}
}

\end{document}